\documentclass[letterpaper, 10 pt, conference]{ieeeconf}  % Comment this line out if you need a4paper
\pdfoutput=1

\IEEEoverridecommandlockouts                              % This command is only needed if 
\usepackage{amsmath} % assumes amsmath package installed
\usepackage{amssymb}  % assumes amsmath package installed
\usepackage{cite}

\usepackage{graphicx}
\usepackage{tikz}
\usepackage{pgfplots}
\usepackage{subcaption}
\usepackage[nolist,nohyperlinks]{acronym}
\usepackage{hyperref}
\usepackage{booktabs}

\makeatletter
\let\NAT@parse\undefined
\makeatother
\usepackage{hyperref}  %hyperref needs to be put at the end!

\usepackage[utf8]{inputenc}

\title{\LARGE \bf
Accurate Mapping and Planning for Autonomous Racing
}

\author{
Leiv Andresen$^{1}$*, 
Adrian Brandemuehl$^{1}$*,
Alex H\"onger$^{1}$*,
Benson Kuan$^{1}$*,
Niclas V\"odisch$^{1}$*,
Hermann Blum$^{1}$, \\
Victor Reijgwart$^{1}$,
Lukas Bernreiter$^{1}$,
Lukas Schaupp$^{1}$,
Jen Jen Chung$^{1}$,
Mathias B\"urki$^{1,3}$,
Martin R. Oswald$^{2}$, \\
Roland Siegwart$^{1}$,
and Abel Gawel$^{1}$% <-this % stops a space
\thanks{* These authors contributed equally to this work}% <-this % stops a space
\thanks{$^{1}$ Autonomous Systems Lab, ETH Zürich. Contact: \href{mailto:leiva@ethz.ch}{leiva@ethz.ch}}%
\thanks{$^{2}$ Computer Vision and Geometry Group, ETH Zürich}%
\thanks{$^{3}$ Sevensense Robotics AG}%
}

\pgfplotsset{compat=1.12}

\begin{document}

\maketitle
\thispagestyle{empty}
\pagestyle{empty}

%%%%%%%%%%%%%%%%%%%%%%%%%%%%%%%%%%%%%%%%%%%%%%%%%%%%%%%%%%%%%%%%%%%%%%%%%%%%%%%%
\begin{abstract}
This paper presents the perception, mapping, and planning pipeline implemented on an autonomous race car. It was developed by the 2019 AMZ driverless team for the Formula Student Germany (FSG) 2019 driverless competition, where it won 1st place overall. The presented solution combines early fusion of camera and LiDAR data, a layered mapping approach, and a planning approach that uses Bayesian filtering to achieve high-speed driving on unknown race tracks while creating accurate maps. We benchmark the method against our team's previous solution, which won FSG 2018, and show improved accuracy when driving at the same speeds. Furthermore, the new pipeline makes it possible to reliably raise the maximum driving speed in unknown environments from 3~m/s to 12~m/s while still mapping with an acceptable RMSE of 0.29~m.
\end{abstract}

%%%%%%%%%%%%%%%%%%%%%%%%%%%%%%%%%%%%%%%%%%%%%%%%%%%%%%%%%%%%%%%%%%%%%%%%%%%%%%%%
\section{INTRODUCTION}
% Alex & Benson
\label{sec:introduction}

% For the past 15 years, competitions in autonomous robotics have been used to push the state of the art algorithms in different robotics fields. In particular, the development of autonomous cars was kickstarted by the DARPA Grand Challenge \cite{darpa_grand_challenge} and continued with the DARPA Urban Challenge \cite{darpa_urban_challenge}.

Autonomous racing has grown in popularity in the past years as a method of pushing the state-of-the-art for various autonomous robots. While competitions generally allow researchers to test the robustness and applicability of different solutions, racing in particular presents additional challenges to computational speed, power consumption, and sensing. 
Autonomous car racing started almost 16 years ago, with the DARPA Grand Challenge \cite{thrun2006stanley}, and now continues with Roborace\footnote{\url{roborace.com}} and \ac{FSD}\footnote{\url{www.formulastudent.de}}.
In addition to car racing, new platforms such as quadcopters have recently begun to pick up momentum in new competitions like the IROS drone racing competition\footnote{\url{ris.skku.edu/home/iros2016racing.html}}.

Two main categories can be distinguished in autonomous racing: racing with and without prior knowledge of the track layout. For instance in Roborace, the race track layout is known a priori, whereas in IROS drone racing, the locations of a set of gates must be discovered during flight. \ac{FSD} offers disciplines from both categories.

\begin{figure}[t]
    \centering
    \includegraphics[width=\linewidth]{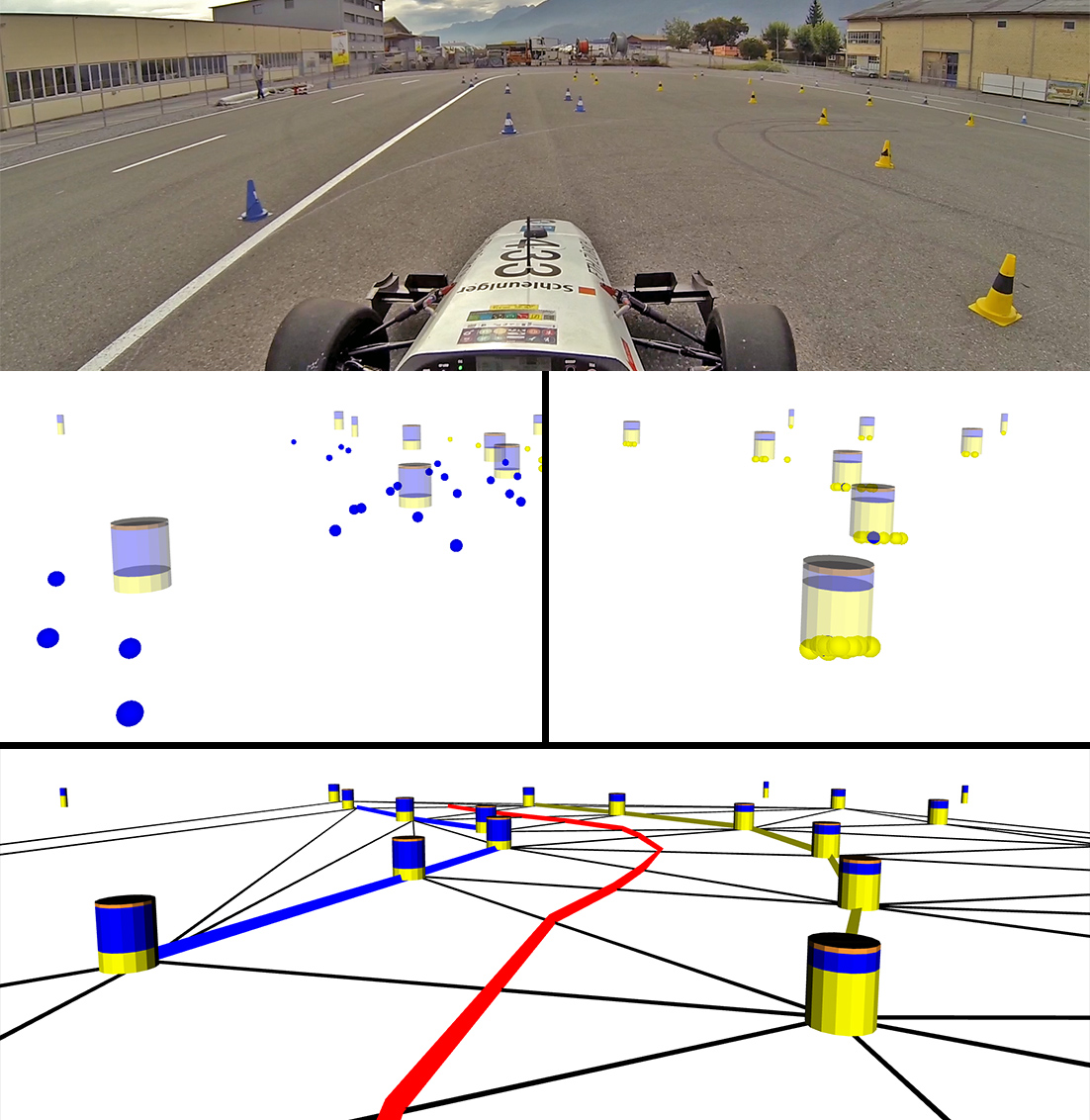}
    \caption{Top: track in the front view of the car. Middle: accumulated observations of cameras (left), LiDARs (right), and filtered estimates (cylinders). Bottom: estimated track by boundaries and middle line.}
    \label{fig:collection}
\end{figure}

Racing on unknown tracks is challenging for SLAM algorithms. First, high-speed motion is more uncertain, which causes higher drift between sensor readings and poses challenges for data association. Furthermore, the horizon over which the upcoming dynamic motion must be planned scales with speed. If a tight corner is detected too late, the robot is unable to slow down in time for the turn, and inevitably leaves the track. This need for a longer horizon presents challenges not only for the planning algorithms but also for perception and mapping. For perception, the density of visual and 3D LiDAR sensor data is proportional to the inverse-square of the distance. Detecting distant landmarks is thus challenging and noise is unavoidable. The latter calls for a mapping pipeline that appropriately models and fuses the uncertainties of different modalities at varying distances. Finally, a planning approach is required that considers the uncertainty of the estimated map. In this work, we present a robotic system that is able to achieve high mapping accuracy while increasing the maximum speed on unknown tracks from 3~m/s \cite{amz_redundant_perception, amz_mapping_system} to 12~m/s. Please refer to Fig.~\ref{fig:collection} and Fig.~\ref{fig:hw_sw_architecture} for an illustration of the pipeline. A video demonstrating our work is available online\footnote{\url{https://youtu.be/1YBFcHY_gWE}}.

Our work is tested in the context of the \ac{FSD} competition, where racing on both unknown and known tracks has been split after the 2018 season. The competition was previously conducted on an unknown ten lap course and autonomous racing pipelines \cite{amz_redundant_perception, amz_mapping_system} focused on the reliability of a usually slow first `mapping lap', knowing that nine fast laps would follow and reduce its effect on the total score. 
With racing in unknown environments as a separate one lap event (\emph{Autocross}), the competition encourages systems to increase driving speed while still providing a high quality map for the consecutive ten lap event (\emph{Trackdrive}) on the same course.

We present a perception, mapping, and planning pipeline designed to minimize lap time on an unknown track while still accurately mapping the track for the \emph{Trackdrive} event.

The contributions of this paper are
\begin{itemize}
    \item a perception pipeline that combines LiDAR and camera readings for accurate early fusion of cone detections,
    \item a two-stage mapping pipeline that allows for planning on lower certainty data while recording a high quality map,
    \item a new path planning algorithm utilizing Bayesian inference to find the track boundaries, and
    \item the integration of the aforementioned methods into a solution capable of planning over long horizons while driving at high speeds.
\end{itemize}

% \begin{itemize}
%   \item (Increase prediction horizon to enable fast driving on unknown track)
%   \item Early Fusion of perception sensor data.
%   \item Comparable map accuracy with high speed driving.
%   \item Bayesian representation for estimation of track boundaries.
% \end{itemize}
\section{Related Work} \label{sec:related_work}

Most related to our work are the Roborace and the \ac{FSD} competitions, both including races on known race tracks, where the emphasis lies on reliable localization and superior control. The authors in \cite{roborace_vehicle_dynamics_localization} and \cite{roborace_lidar_localization} deal with localization in known track environments, using only LiDAR and GNSS in combination with open-source ROS packages such as AMCL. Our previous works~\cite{amz_mapping_system, amz_redundant_perception, amz_paper_2019} also address this challenge by improving perceptual robustness with a novel detector and probabilistic sensor fusion scheme for cone observations on race tracks.

\ac{FSD} also includes racing on a previously unknown race track, which became a separate discipline in its latest edition. Racing on an unknown track pushes requirements on detection range as it directly affects the achievable maximum speed. This is in addition to existing requirements on map quality as in previous AMZ works~\cite{amz_mapping_system, amz_redundant_perception, amz_paper_2019} and other \ac{FSD} teams \cite{strobelaccurate}.

A key task of the presented perception system is the detection of cones. In \cite{strobelaccurate}, and \cite{de2019cnn} the authors present state-of-the-art CNN-based vision-only detection systems for cone detection based on versions of YOLO \cite{redmon2017yolo9000, redmon2018yolov3}. In order to make better use of the accurate range measurements from LiDAR in combination with the rich semantic information from cameras, we implement an early sensor fusion approach for accurate color and position detection of cones.
State-of-the-art object detectors that fuse camera and LiDAR data \cite{avod, frustrum_pointnets} typically require powerful GPUs to run in real-time, making them unsuitable for racing systems where space and packaging for power supplies are challenging. We achieve real-time performance on a mobile version of the GTX 1070 without hitting the upper limit of our computing system.

We combine early fusion for cone detection with a Kalman Filter based local map. This yields improved cone localization and color estimation performance, by accumulating and filtering the measurements over time. For redundancy in case of sensor failures, vision-only and LiDAR-only measurements can still be incorporated in similar fashion to our previous work \cite{amz_redundant_perception}. The authors in \cite{vangool2010} develop a method for tracking dynamic obstacles, including discrete object categories, using late fusion via an Extended Kalman Filter. Similarly, \cite{chen2012} implements the fusion as an \ac{UKF}. Since the landmarks that we wish to estimate are static, no nonlinear update terms are needed and we instead rely on plain Kalman Filters.

Once a partial map is observed, planning in partially known environments is well studied. Continuous state and action space planning methods such as RRT* and PRM* \cite{sampling_motion_planning} are computationally complex, and do not exploit the specific patterns of the tracks that we aim to follow. Brandes et. al.~\cite{mitlanedetection} uses binary integer optimization on a fixed number of cones to find the immediate boundaries of a Formula Student track, however due to the low number of cones used it cannot give boundaries far enough from the vehicle to drive at high speeds safely. We develop an algorithm that is both computationally efficient and exploits the nature of the planned paths for highly accurate results with dynamic numbers of environment landmarks.

\section{METHOD}
\subsection{Overview of the Autonomous Driving Pipeline}
\label{sec:method_overview}

% Notation:
% \begin{itemize}
%     \item \textbf{Car pose:} \\  ground truth = $x^t = \textcolor{red}{\{\mu_x, \mu_y, \mu_{\theta}, \sigma\}}$, \\  estimation = $\hat{x}^t$, since it is estimated by Velocity estimation and integrated. 
%     \item \textbf{Cone $i$'s position at time $t$:} \\ ground truth = $z^{t}_i = \textcolor{red}{\{\mu_x, \mu_y, \sigma\}}$, \\ perception estimate = $\hat{z}^{t}_i$, \\ local map filtered estimate = $\tilde{z}^{t}_i$, \\ Cone location arrays = $Z^t$, $\hat{Z}^t$, $\tilde{Z}^t$
%     \item \textbf{Cone $i$'s color at time $t$:} \\ ground truth = $c^{t}_i = \textcolor{red}{\{p_b, p_y, p_o\}}$, \\ perception estimate = $\hat{c}^{t}_i$, \\ local map filtered estimate = $\tilde{c}^{t}_i$, \\ Cone color arrays = $C^t$, $\hat{C}^t$, $\tilde{C}^t$
% \end{itemize}{}

In \ac{FSD}, tracks are demarcated by yellow, blue, and orange traffic cones (see Fig. \ref{fig:collection}). Yellow and blue cones mark the right and left boundaries, respectively, while orange cones mark the start line of the lap. In \emph{Autocross}, track limits such as min/max cone separation, total track length, etc. are given by the rules \cite{fsg_rules}, but the track is otherwise unknown.

\begin{figure}[b]
     \centering
     \includegraphics[width=\linewidth]{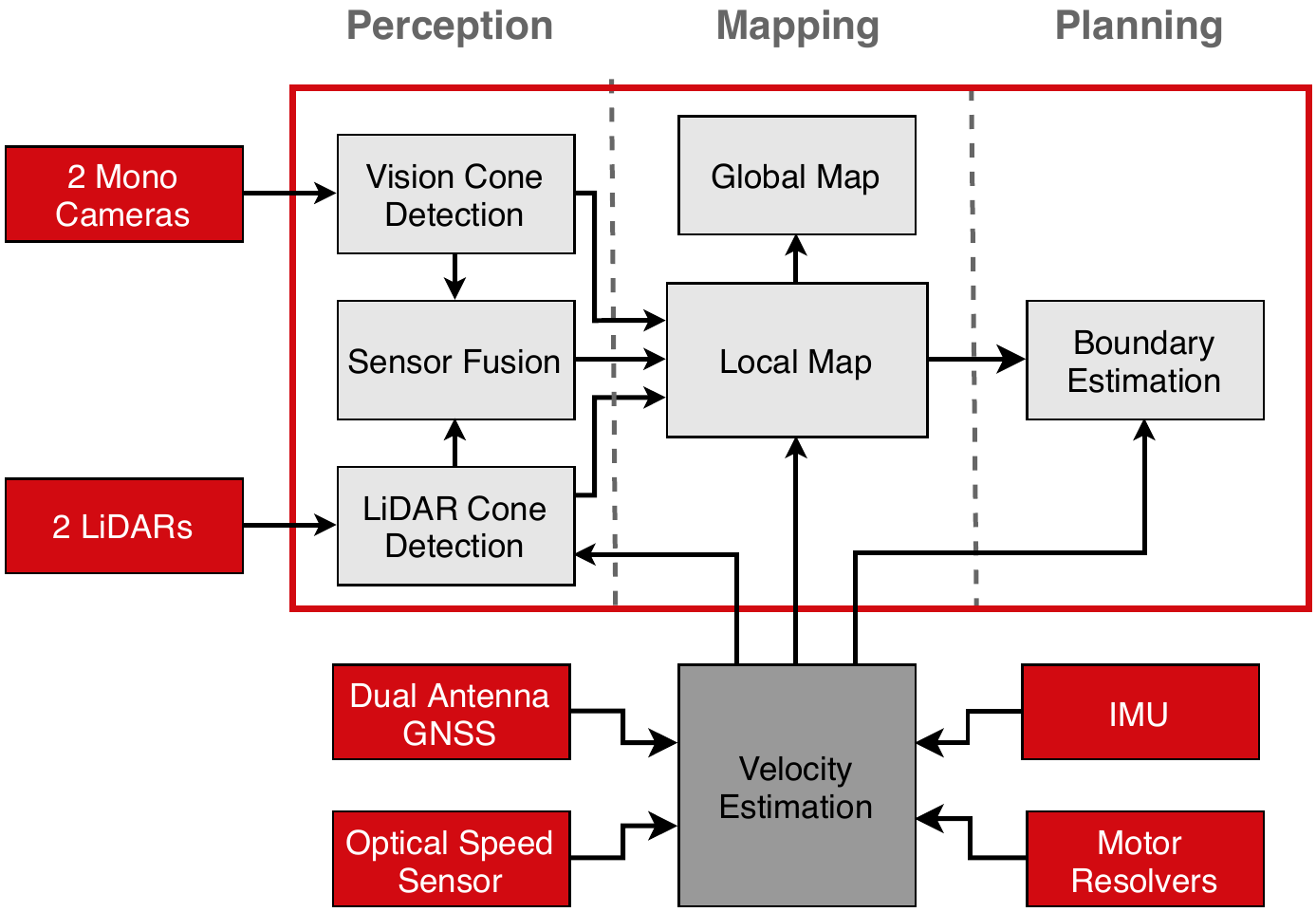}
     \caption{Architecture of the autonomous driving pipeline.}
     \label{fig:hw_sw_architecture}
\end{figure}

Fig. \ref{fig:hw_sw_architecture} depicts the architecture of the autonomous driving pipeline we developed. The maximum speed of the car on an unknown track is directly limited by the distance at which the track boundaries can be reliably identified. Thus, the perception setup is designed to detect cones over a long range and with a large horizontal field of view. It comprises two 32 channel rotating LiDARs and two monocular cameras with global shutter; all devices are time synchronized to a common GNSS clock. To achieve better cone detection performance an early fusion approach that combines these two modalities was used. 
The longitudinal, lateral, and rotational velocities $\hat{v}^t=\{v_x,v_y,\psi\}$ of the car are estimated for ego-motion compensation and mapping. The estimate is obtained by fusing raw data from multiple sources including an Inertial Measurement Unit, dual antenna GNSS, an optical speed sensor, motor resolvers, and the motor currents with an Unscented Kalman Filter \cite{julier1997}. 
Utilizing the cones as landmarks, a local map fuses velocity estimates and the cone observations from the perception pipelines over time. The local map is used to estimate the track boundaries and middle path for the \emph{Autocross} discipline with Delaunay triangulation and Bayesian inference. A globally consistent map is estimated using GraphSLAM to represent the cone positions accurately in the world frame for use in the later \emph{Trackdrive} event. The definition of local and global mapping is further explained below in section \ref{sec:method_mapping}.

\subsection{Perception}
\label{sec:method_perception}

Position and color of cones can be detected independently by the camera and LiDAR pipelines. An optional early sensor fusion approach combines both modalities to improve the overall accuracy and robustness.

\subsubsection{Camera only}
The vision pipeline utilizes a convolutional neural network (CNN), Tiny YOLOv3 \cite{redmon2018yolov3}, that is trained to detect blue and yellow cones in the images of the mono cameras. Based on the size of the bounding boxes and the true dimension of a cone, the distance to the cone is estimated and the detected cones are projected into 3D space accordingly.

\subsubsection{LiDAR only}
The LiDAR pipeline consists of three stages: 1) pre-processing, 2) cone detection, and 3) cone classification. This pipeline follows the method described in \cite{amz_redundant_perception}, which  presents an approach to estimate the color of a cone using only LiDAR point cloud data. A neural network predicts the most probable color based on the pattern of LiDAR intensity returns across multiple channels, e.g. a yellow cone is characterized by a bright-dark-bright pattern.

\begin{figure}[t]
    \centering
    \includegraphics[width=\linewidth]{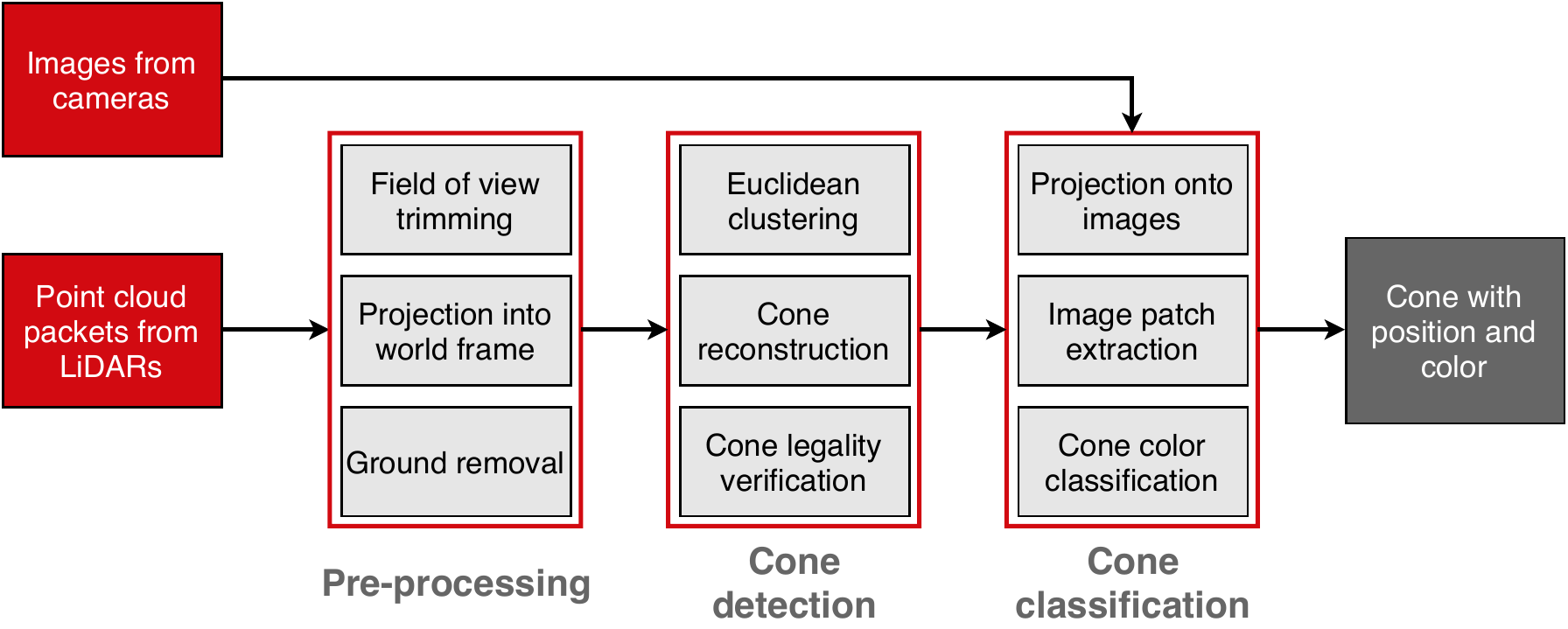}
    \caption{The LiDAR pipeline used to detect cones from point clouds and to classify them according to their color via image patches. It consists of three major phases: pre-processing and cone detection following \cite{amz_redundant_perception}, cone classification is implemented by our early sensor fusion approach.}
    \label{fig:lidar_pipeline}
\end{figure}

\subsubsection{Sensor Fusion}
\begin{figure}[b]
    \centering
    \includegraphics[width=\linewidth]{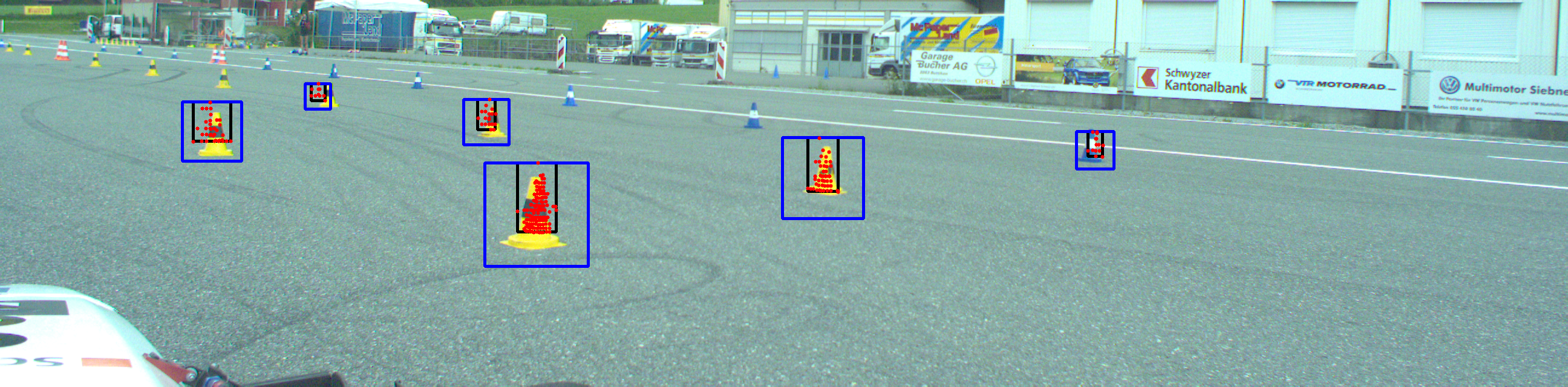}
    \caption{Projection of the cones' point clouds onto both image planes. The points are shown in red, the initial bounding boxes are black, and the final image patches are indicated by blue squares.}
    \label{fig:lidar_projection}
\end{figure}

Both the vision and LiDAR pipelines can run independently and therefore make the system robust to failures of single sensors, as was described in~\cite{amz_redundant_perception}. In the case that both sensor modalities are available, our sensor fusion approach on the input data level exploits the respective advantage of each modality. The overall pipeline is shown in Fig. \ref{fig:lidar_pipeline}. Once the positions of the cones have been determined using the aforementioned LiDAR-based cone detection, the points of each cone are projected onto the image of either camera depending on their lateral position. Since every point contains a timestamp, it is possible to interpolate them to the exact time of the image using the velocity estimates. The projected points of each cone are then framed by bounding boxes, as shown in Fig. \ref{fig:lidar_projection}. Since points on the lower part of a cone are often missing due to ground removal, the height of the initial boxes is multiplied by 1.5, which worked best in our experiments. Afterwards, the boxes' width is set to match their height. The resulting squares are scaled to a uniform size of $32\times32$ pixels, for subsequent classification.

\begin{figure}[b]
    \centering
    \includegraphics[width=\linewidth]{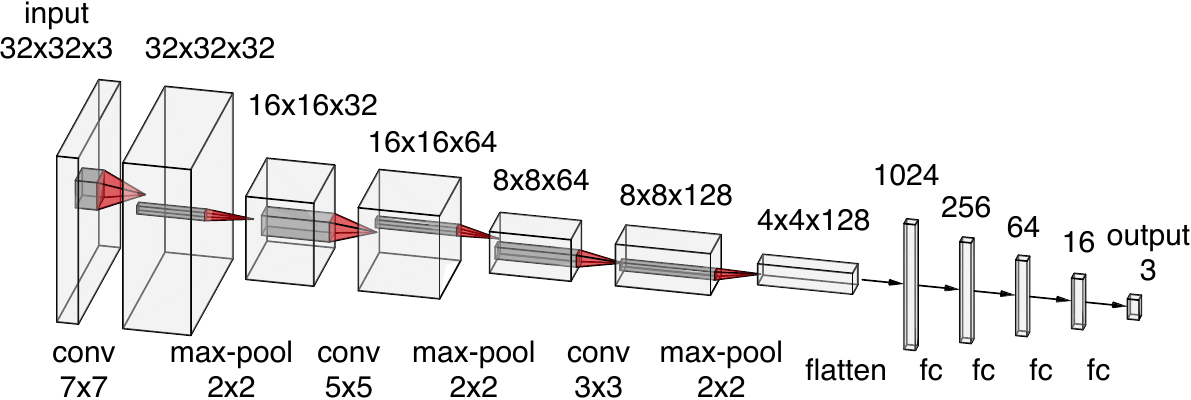}
    \caption{CNN used to predict the cones' color based on RGB image patches.}
    \label{fig:lidar_network}
\end{figure}

Finally, the RGB patches are fed through a customized CNN, see Fig. \ref{fig:lidar_network}, which is inspired by \cite{amz_redundant_perception}. It consists of three convolutional layers, where each is combined with batch normalization, dropout, and a max-pooling layer, as well as four fully connected layers. Softmax is used to compute probability values for the three available classes: blue cone, yellow cone, and unknown color. To deal with occlusion, the training dataset contains samples of multiple cones of both mixed and the same colors within a single image patch, where the latter case is less critical for cone color detection.

% \begin{figure}[b]
%     \centering
%     \includegraphics[width=\linewidth]{figures/lidar_pipeline.pdf}
%     \caption{The LiDAR pipeline used to detect cones from point clouds and to classify them according to their color via image patches. It consists of three major phases: pre-processing and cone detection following \cite{amz_redundant_perception}, cone classification is implemented by our early sensor fusion approach.}
%     \label{fig:lidar_pipeline}
% \end{figure}

\subsection{Mapping}
\label{sec:method_mapping}

The three perception pipelines are filtered and fused together into a local map. In a second step, the output of this local map is used to build a global map of the whole track. The local and global map are detailed in the following:

\subsubsection{Local map}
\label{sec:local_map}

Redundancy against sensor pipeline failures requires a mapping approach that works for all combinations of functioning pipeline subsets. Single sensor pipelines prove to be highly uncertain in either cone depth (monocular vision) or cone color estimation (LiDAR). Furthermore, all sensor pipelines suffer from significant noise when detecting distant cones. The global map requires observations to be sufficiently certain in order to avoid damage stemming from data association errors. On the other hand, path planning also requires knowledge of far and uncertain cones in order to evaluate sufficiently long planning horizons for fast driving. These two requirements can be satisfied by using a local map that estimates the position and color probabilities for all cones that were recently observed.

Sensor pipeline failures are detected by a time threshold and parameters are adjusted to one of four modes of operation. When the early sensor fusion pipeline is operational, no other pipelines are used to update the local map since they would fuse in redundant information. When the early fusion pipeline is not operational, the strengths of the individual modalities are modeled with low covariances on the cone positions from the LiDARs and cone colors from the cameras. Two modes of operation remain, which correspond to individual failures of the camera or LiDAR pipeline. Under pipeline failures, the goal shifts towards reliably finishing the lap, in order to still generate a map for the \emph{Trackdrive} event. A speed of 5 m/s is tested and verified for all modes of operation.

A feature-based single hypothesis map $\boldsymbol{\mathcal{M}}_{\rm local}$ is chosen, since it is sufficient to accurately represent the cones that were recently observed. The pose estimate $\hat{\boldsymbol{x}}^t = \{\mu_x, \mu_y, \mu_{\theta}\}$ of the car within $\boldsymbol{\mathcal{M}}_{\rm local}$ is computed by integration of the estimated velocity $\hat{\boldsymbol{v}}^t$, starting at $\hat{\boldsymbol{x}}^0 = \{0, 0, 0\}$.
New cone observations $\hat{\boldsymbol{z}}^{t}$ are transformed from car frame to $\boldsymbol{\mathcal{M}}_{\rm local}$ using the synchronized sensor timestamps. 
%New observations $\hat{z}^{t-\Delta delay}$ are transformed from car frame to $\mathcal{M}_{\rm local}$ with respect to $\hat{x}^{t-\Delta delay}$, thus compensating for time delay. 

Cone positions and uncertainties are filtered using a Kalman Filter. 
The colors of the $n$ observed cones at time $t$ are represented by the random variable $\hat{\pmb{C}}^t=[\hat{C}_1^t,\ldots,\hat{C}_n^t]$. The normalized sum of the probability distribution of $\hat{\pmb{C}}^t$ over time yields a filtered color estimate up to time $t$. This filtered output is represented as $\tilde{\pmb{C}}^t=[\tilde{C}_1^t,\ldots,\tilde{C}_n^t]$, where each cone $i$ takes one of the following classes, $\tilde{C}_i^t=\{\rm yellow, blue, unknown\}$. The motion uncertainty is incorporated by growing the pose covariance of all cones in $\boldsymbol{\mathcal{M}}_{\rm local}$ at each time step, using a hand tuned odometry covariance.

Associations of cone detections between time steps are challenging, as the noise levels on observed cone positions often exceed the minimal expected distance between neighboring cones. Each cone observation $\hat{\boldsymbol{z}}$ is therefore associated to its closest cone in $\boldsymbol{\mathcal{M}}_{\rm local}$ using the Bhattacharyya distance \cite{bhattacharyya}, which incorporates the position covariance of both the observation and the cone estimates in the map.

Furthermore, a negative observation model is used to reject non-existent cones resulting from false positives or outliers within less than 0.5s even for previously highly certain false positive cones. This is achieved by penalizing the certainty of cones in $\boldsymbol{\mathcal{M}}_{\rm local}$ that are not observed despite being in the field of view of the sensors.
The parameters of the observation model are derived from ground truth data analysis of the observation recall rates. The observation noise is also computed from the ground truth data. 
%Other mapping parameters are optimized by random search. An RMSE-based error metric is used, which takes the cone recall and false-positive rates into account.

An output $\boldsymbol{\mathcal{M}}_{\rm local}^{t_n}$ containing all cones within the map is created for every set of new observations measured at time $t_n$ to be used by the global map and planning.

%\begin{itemize}
%  \item Explain the need of filtering perception
%  \item Explain the map representation and reasons (single hypotheses, no loop closures)
%  \item How is data associated? Why Bhattacharyya? Delay handling?
%  \item Explain how cones are updated shortly
%  \item Concept of negative weighting based on sensor models
%  \item Pipeline-specific output
%  \item How can ground truth data be used to improve mapping?
%  \item Random search usage for parameter optimization
%
%\end{itemize}

\subsubsection{Global map}

To drive multiple laps as fast as possible in the \emph{Trackdrive} event, a global map containing all cones is generated and used for localization and racing line optimization. The map must be accurate to allow downstream algorithms to plan trajectories close to the cone delimited track boundaries.

The global map is generated in real-time on the car using a pose-landmark GraphSLAM algorithm \cite{article_GraphSLAM_toturial}. Cones are used as landmarks, and the car and cone poses are jointly estimated via non-linear least squares using Google's Ceres solver \cite{ceres-solver}.

Cone pose estimates are added to the (global) graph as virtual measurements from every new local map $\boldsymbol{\mathcal{M}}_{\rm local}^{t_n}$. In order to reduce the chance of false associations and outliers, new cones are only added to the global map if they were observed in the last perception reading and are in close proximity to the car. For as long as cones are in the local map, the data associations from $\boldsymbol{\mathcal{M}}_{\rm local}^{t_n}$ are re-used. Cones that are new in the local map are associated to landmarks in the graph using euclidean distance. If there is no landmark within a certain distance threshold, a new landmark is created. Graph edges connecting car poses are given by the integration of the velocity estimate between two subsequent local maps $\boldsymbol{\mathcal{M}}_{\rm local}^{t_n}$, $\boldsymbol{\mathcal{M}}_{\rm local}^{t_{n-1}}$.

Directly using the vision and LiDAR measurement models in the graph's factors would improve accuracy. However, we primarily use GraphSLAM to distribute the accumulated pose error when revisiting the finish line over the entire trajectory and map. Improving the measurement models and incorporating explicit place recognition will be the subject of future work.

\subsection{Planning}
\label{sec:method_planning}

In the \emph{Autocross} discipline, the car must drive around an unknown track. Using the filtered cone observations from the local map as inputs, we estimate the track boundaries and middle path online in three steps. For the first and second steps, our approach follows the previous approach described in the Boundary Estimation section of \cite{amz_paper_2019}. In the first step, the search space is discretized using the Delaunay triangulation and the second step generates candidate middle paths from the triangulation.
% The environment is discretized into a graph with Delaunay triangulation, using the cones' position as vertices to form triangles. A tree of possible middle paths is then generated in this graph from a single starting point by expanding to the mid-points of neighbouring triangulation edges. The vertices of the triangulation edges that were expanded to, became the corresponding boundary points for the middle path. 
Our approach differs in the third step: it uses Bayesian inference instead of a single cost function to select the best middle path and its corresponding boundary points. The use of Bayesian inference allows the cone color probabilities to have a better representation of the validity of the path. This will be explained further in Section~\ref{sec:likelihood}.
% As the first two steps are similar to the one in \cite{amz_paper_2019}, 

The normalization term of the Bayesian inference is omitted as it does not affect the final selection result. Thus, our inference equation consists of only three terms: the posterior $Pr({\rm valid\,path}|\,\tilde{\pmb{C}}^t)$, the prior $Pr({\rm valid\,path})$, and the likelihood $Pr(\tilde{\pmb{C}}^t|{\rm valid\,path})$. 
% These three terms will be explained in greater detail in the following subsections below.

% \begin{equation} \label{eq:bayesinfer}
% \begin{split}
% &Pr(valid\,path|cone\,color) =\\
% &\phantom{{}---}Pr(cone\,color|valid\,path) \cdot Pr(valid\,path) 
% \end{split}
% \end{equation}

\subsubsection{Prior}
\label{sec:prior}
The prior is the probability of a cone configuration representing a valid path. The validity of a path is determined through human experience, which is captured in the form of six hand-crafted features. The function to calculate the prior value, consists of a prior weight $W_{prior}$ and for each feature instance $j$, a feature value $F_j$, scaling factor $N_j$, setpoint value $S_j$, and feature weight $W_{j}$. It has the following form:

\begin{equation}
{\rm Prior} = Pr({\rm valid\,path}) = e^{-W_{prior}\cdot {\rm Cost}},
\end{equation}

\begin{equation}
{\rm Cost} = \sum_{j=1}^6\frac{W_{j}\cdot(F_j-S_j)^2}{N_j}.
\end{equation}

The hand-crafted features, $F_j$, are defined as follows: $F_1$ for the largest angle change within the path; $F_2$, $F_3$ and $F_4$ for the standard deviation of the distance between the left cones, the right cones and the track width respectively; $F_5$ for the number of triangulation edges crossed limited to the desired number; and $F_6$ for the length of the path.
 
\subsubsection{Likelihood}
\label{sec:likelihood}

The likelihood is the product of the color probability of the cones that are required to estimate the middle path from the prior. The color probability of a cone is distributed between the classes blue, yellow, and unknown. Cones on the left boundary can be either blue or unknown and will be assigned the higher probability value of either color. The same applies to cones on the right boundary which are either yellow or unknown. Cones that are not boundary points of the path will be assigned the highest probabilistic value amongst all three colors. 
% An example of how probability values are assigned to cones given a middle path is shown in Figure \ref{fig:likelihood}. In this example, every cone is assumed to have 0 probability value for the color orange. The assigned probabilistic value for each cone is then multiplied together to give the likelihood. 
The likelihood when $n$ cones are detected, and the assigned color of each cone being represented by $c_i$, has the following form,

\begin{equation}
\begin{split}
{\rm Likelihood} &= Pr(\tilde{\pmb{C}}^t|{\rm valid\,path}) \\
           &= \prod_{i=1}^n Pr(\tilde{C}_i^t = c_i).
\end{split}
\end{equation}

With this probabilistic approach for the likelihood, the color probabilities of every observed cone are used to calculate the validity for each candidate path. As compared to the previous approach \cite{amz_paper_2019}, where only the color probabilities of cones belonging to the boundaries of a candidate path are considered. This results in the color probability information of the other cones to be ignored. 
% Low order statistics also may not be the best way to compress information from different number of cones into a similar form for comparison. Whereas the multiplication of probabilities of independent cone color together is an established approach to obtain a single probabilistic value for any combination of colors assigned to each of the cones. 

\subsubsection{Posterior}
\label{sec:posterior}

The aim of obtaining the posterior value for each possible middle path is for comparison, so that the middle path with the highest posterior value can be selected. 
% Since removal of the evidence term from the calculation of the posterior does not change the ranking of the middle paths, it is removed to reduce computation. 
To ensure numerical stability, $\log$ is applied to the new posterior. The formulation for the new posterior is now a sum of the $\log$ of the prior in section \ref{sec:prior} and the $\log$ of the likelihood in section \ref{sec:likelihood}. This formulation is shown in the equation below,

\begin{equation}
\begin{split}
{\rm \log(Posterior)} &= \log(Pr({\rm valid\,path}|\,\tilde{\pmb{C}}^t)) \\
          &\propto \log({\rm Prior}) + \log({\rm Likelihood})\\
          &= -W_{prior}\cdot {\rm Cost} + \sum_{i=1}^n \log(Pr(\tilde{C}_i^t = c_i)).
\end{split}
\end{equation}

\section{IMPLEMENTATION}
% platform (mechanical/eletrical(computer)), Experiment setup/data
\subsubsection{Robotic platform} 

The methods presented in this paper were implemented and tested on the autonomous race car \emph{pilatus driverless}, a newer base vehicle compared to \cite{amz_redundant_perception}. It is also powered by four electric motors and offers a full aerodynamics package.

\begin{figure}[h]
    \centering
    \includegraphics[width=\linewidth]{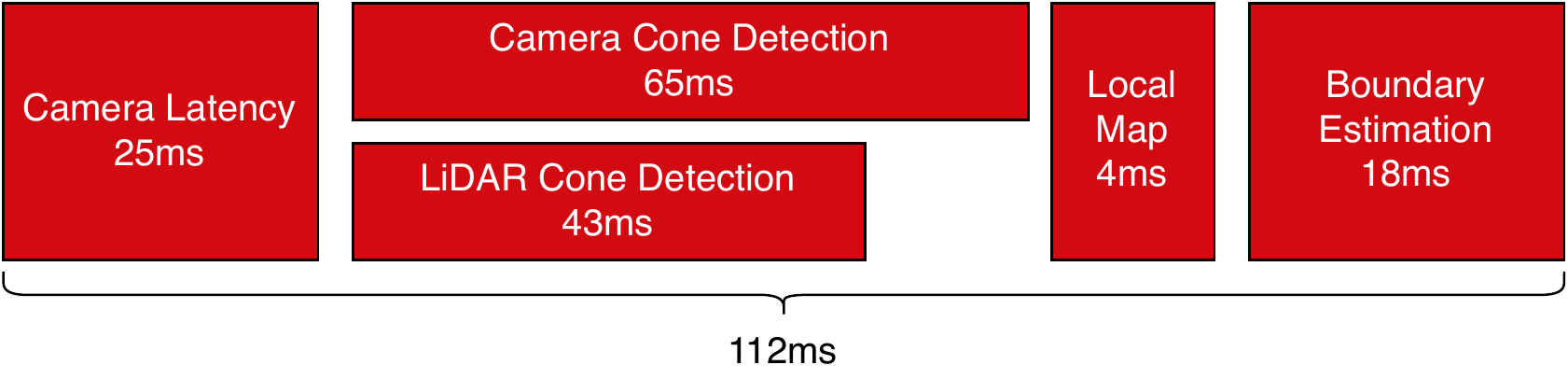}
    \caption{Processing times of each pipeline component.}
    \label{fig:latency}
\end{figure}

All information is processed by a quad-core Intel Xeon E3-1505M with 16~GB of RAM using ROS and a real-time capable electronic control unit. The actuators include an electric power steering system, four self-developed wheel hub motors with regenerative braking, and an emergency brake system. 
Latencies for all pipeline elements can be found in Figure \ref{fig:latency}.

%\begin{itemize}
%  \item dynamic qualities
%  \item computing hardware
%  \item Calibration and Time Synchronization
%\end{itemize}

\subsubsection{RTK ground truth data}

Results in this paper are generated by comparison to ground truth data. Ground truth maps and trajectories are measured through GNSS with Real-Time Kinematic (RTK) corrections. A dual antenna GNSS RTK system installed on the car provides ground truth position and heading. A hand-held device with a second GNSS RTK system and user interface enables measuring cone locations and colors, and generates ground truth maps of the track with up to 2~cm accuracy for each cone. While usage of RTK is allowed in the competition, no guarantee of base station availability is made. Therefore, it is not used in the autonomous pipeline.

\subsubsection{Experimental setup}

Most experiments are conducted on \emph{pilatus driverless}, the aforementioned race car, on various tracks at competitions and test sites. All tracks contain challenging segments, i.e. hairpins and parts of the track that come close to each other. The tracks' lengths vary between 200~m and 300~m and the race car's speed is predominantly in the range of 5-12~m/s. The results are compared to \cite{amz_redundant_perception}. Where mentioned, some experiments use our approach on sensor data from \cite{amz_redundant_perception} to account for the differences in vehicle and sensor setup.

\section{RESULTS}
% 3 main experiments
% \begin{itemize}
%     \item BASELINE: gotthard data (wemding rosbag) and code, 
%     \item OUR APPROACH: pilatus data (post season rosbag) and code
%     \item OUR APPROACH NORMALIZED FOR SENSOR SETUP: gotthard data (wemding rosbag) and pilatus code.
% \end{itemize}{}
% Result METRICS: Histogram of erroneous path over distance, map accuracy RMSE

\subsection{Perception}
\label{sec:results_perception}

We compared our novel low-level sensor fusion approach for cone color classification to the LiDAR-only method in \cite{amz_redundant_perception}. To account for the improved sensor setup, we re-trained this network with new data recordings. 

Cameras capture richer color information than LiDAR and, hence, enable more robust cone classification. By artificially enlarging the area of interest around the projected points in the image, the method becomes more robust to imperfect projection due to errors in calibration and ego-motion estimation. Especially for larger distances, our image-based classification benefits from the rich representation. During the training of both networks, we only utilize data within a 10~m range. The results in Table \ref{tab:lidar_comparison_distance} show that the performance of the intensity pattern-based method drops above 12.5~m, while the performance of our method does not change. Such an improvement enables higher certainty in the detection of the track and, ultimately, faster racing.

\begin{table}[h]
\begin{center}
        \caption{Accuracy of color classification of correctly detected cones with respect to distance.}\vspace{1ex}
        \label{tab:lidar_comparison_distance}
        \setlength{\tabcolsep}{5.5pt}
        \begin{tabular}{lccccc}
        \toprule
        % & \multicolumn{5}{c}{Accuracy} \\
        %\hline
        Range [m] & 0-5 & 5-7.5 & 7.5-10 & 10-12.5 & 12.5-15 \\
        %\hline
        \midrule
        previous approach \cite{amz_redundant_perception} & 0.88 & 0.93 & 0.89 & 0.87 & 0.80 \\
        our method & 0.99 & 0.99 & 1.00 & 1.00 & 1.00 \\
        \bottomrule
        \end{tabular}
\end{center}
\end{table}
\subsection{Mapping}
\label{sec:results_mapping}

% \subsubsection{Local}
% \begin{itemize}
%   \item BE-metrics with and without improved parameters by GT/Param opt
%   \item Maybe error metrics of LM to compare accuracy in different sensor modes
% \end{itemize}

% \subsubsection{Global}
We evaluate the accuracy and correctness of the global map generated by our mapping pipeline by aligning the estimated and ground truth maps with ICP \cite{121791}, \cite{article_ICP_PTP}. Then we compute the RMSE of the estimated cone locations with respect to their ground truth locations. We compare our results to the approach described in \cite{amz_redundant_perception}. To compensate differences in the sensor setup, we applied our method to the same input sensor data used in \cite{amz_redundant_perception}. Finally, we show the accuracy of our method on the improved sensor setup. The results are shown in Table~ \ref{tab:mapping_accuracy_comparison}.

\begin{table}
\begin{center}
        \caption{Comparison of mapping accuracy (RMSE) with various speeds and methods}\vspace{1ex}
        \label{tab:mapping_accuracy_comparison}
        \begin{tabular}{llcc}
        \toprule
        Method & Dataset & Max. mapping speed & RMSE \\
        \midrule
        previous approach \cite{amz_redundant_perception} & 2018 \cite{amz_redundant_perception} & 2.8 m/s & 0.20 m \\
        our method & 2018 \cite{amz_redundant_perception} & 2.8 m/s & 0.16 m \\
        our method & 2019 & 12 m/s & 0.29 m \\
        \bottomrule
        \end{tabular}
\end{center}
\end{table}

\begin{figure}[t]
    \centering
    \includegraphics[width=\linewidth]{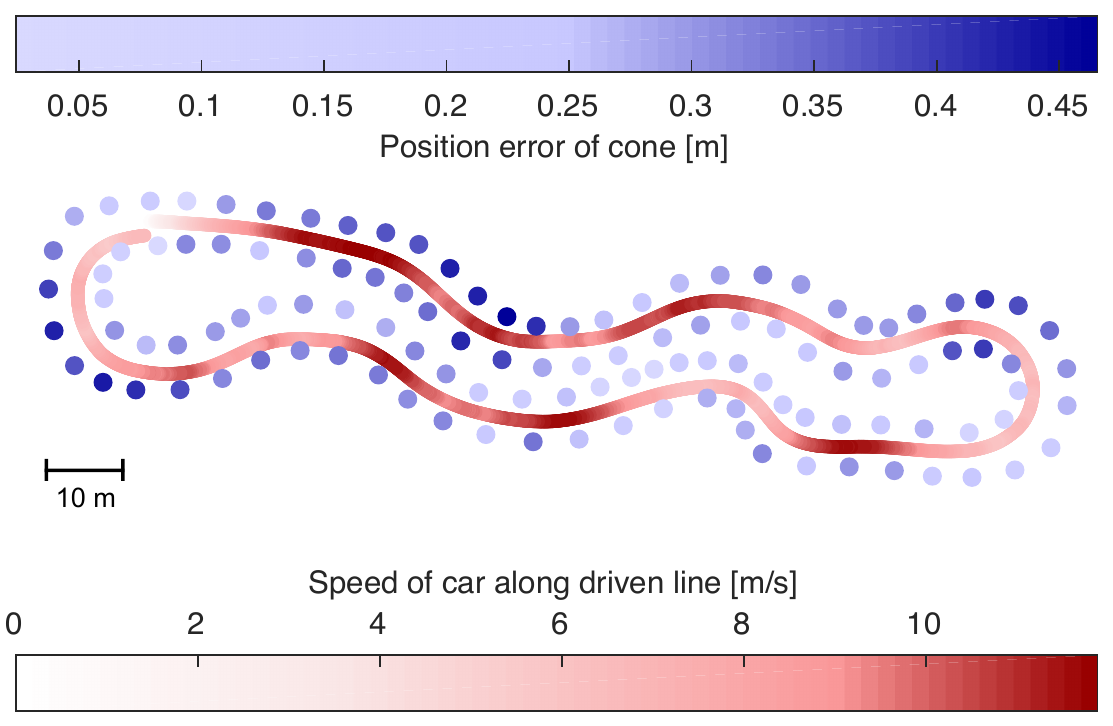}
    \caption{Position error of the estimated cone locations relative to the ground truth. The driven line is colored according to the speed of the car.}
    \label{fig:mapping_vs_gt_tuggen}
\end{figure}

These results show that our method allows for more than a 4-fold increase in maximum mapping speed with a moderate decrease in mapping accuracy. Additionally, we show that our method achieves superior accuracy to the previous implementation when used on the same perception input data. Fig. \ref{fig:mapping_vs_gt_tuggen} shows the mapping performance of our method for a 213~m long track recorded at a test site. 

% \begin{itemize}
%   \item Data Association Experiment Setup: feature size, how metrics where generated
%   \item Data Association performance: \% correct, \% no prediction
%   \item (Data Association performance w. simulated state estimation noise) 
%   \item Global Map accuracy vs. ground truth: Before and after optimization
%   \item Global Map accuracy vs. previous implementation: FastSLAM from old paper
% \end{itemize}
\subsection{Planning}
\label{sec:results_planning}

Using the parameter values, 
$W_k=0.1 \mid k \in \mathbb {Z} : 1\leq k \leq 5$, 
$W_6 = 0.5$ and $W_{prior} = 29$ for the Bayesian inference approach, we compare our implemented system against \cite{amz_paper_2019} using new data and recorded perception data from \cite{amz_redundant_perception}.
First, we evaluate our local map from section \ref{sec:local_map} and boundary estimator from section \ref{sec:method_planning} using the perception data from our previous work\cite{amz_paper_2019}. Second, we compare our entire pipeline using the data from our \emph{Autocross} run in the FSG~2019 competition. 

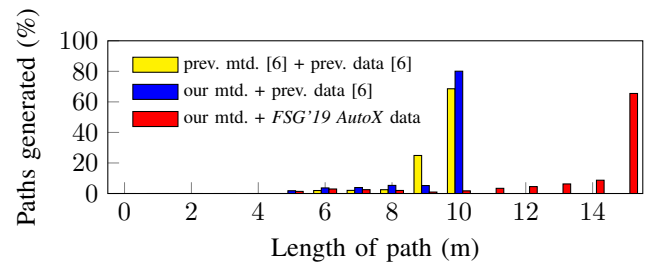
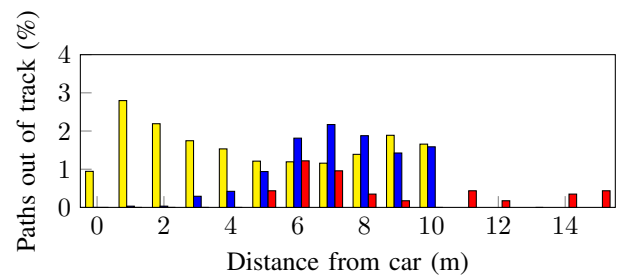
\begin{figure}[h]
    \centering
    
    \begin{subfigure}[b]{0.5\textwidth}

        \begin{tikzpicture}
    
            \begin{axis}[%
            scale only axis,
            width=2.8in,
            height=0.8in,
            legend cell align = left,
            legend style={font=\scriptsize, draw=none},
            legend pos=north west,
            xlabel={Length of path (m)},
            ylabel={Paths generated (\%)},
            xtick pos=left,
            ytick pos=left,
            xmin=-0.5, xmax=15.5,
            ymin=0, ymax=100]
            \addplot[
              ybar,
              bar width=0.04in, 
              bar shift=-0.04in,
              area legend,
              fill=yellow,
              draw=black] 
              plot coordinates{ 
                (0,0) (1,0) (2,0) (3,0) (4,0) (5,0) (6,1.95973632638518) (7,2.04881525031178) (8,2.49420986994477) (9,24.9064671298771) (10,68.5907714234812)
              };
              \addplot[
              ybar,
              bar width=0.04in, 
              bar shift=0.0in,
              area legend,
              fill=blue,
              draw=black] 
              plot coordinates{ 
                (0,0) (1,0) (2,0) (3,0) (4,0) (5,1.81347150259067) (6,3.65932642487047) (7,3.91839378238342) (8,5.31088082901554) (9,5.21373056994819) (10,80.0841968911917)
              };
              \addplot[
              ybar,
              bar width=0.04in, 
              bar shift=0.04in,
              area legend,
              fill=red,
              draw=black] 
              plot coordinates{ 
                (0,0) (1,0) (2,0) (3,0) (4,0) (5,1.39372822299652) (6,2.9616724738676) (7,2.52613240418118) (8,2.00348432055749) (9,0.958188153310104) (10,1.74216027874564) (11,3.39721254355401) (12,4.52961672473868) (13,6.27177700348432) (14,8.71080139372822) (15,65.5052264808362)
              };
            \legend{prev. mtd.\cite{amz_paper_2019} + prev. data\cite{amz_paper_2019},
            our mtd. + prev. data\cite{amz_paper_2019},
            our mtd. + \emph{FSG'19 AutoX} data}
            \end{axis}
        \end{tikzpicture}
        
         \caption{Normalized histogram of percentage of tracks generated for different distance intervals from the car.}
        \label{fig:hist_pred_dist}
    
    \end{subfigure}
    
    \begin{subfigure}[b]{0.5\textwidth}
    
    \begin{tikzpicture}

        \begin{axis}[%
        scale only axis,
        width=2.8in,
        height=0.8in,
        xlabel={Distance from car (m)},
        ylabel={Paths out of track (\%)},
        xtick pos=left,
        ytick pos=left,
        xmin=-0.5, xmax=15.5,
        ymin=0, ymax=4]
        \addplot[
          ybar,
          bar width=0.04in, 
          bar shift=-0.04in,
          fill=yellow,
          draw=black] 
          plot coordinates{ 
            (0,0.944236593621949) (1,2.79707821129521) (2,2.19134152859433) (3,1.74594690896134) (4,1.5321574915375) (5,1.21147336540175) (6,1.19365758061643) (7,1.15802601104579) (8,1.38963121325494) (9,1.8884731872439) (10,1.65686798503474)
          };
          \addplot[
          ybar,
          bar width=0.04in, 
          bar shift=0.0in,
          fill=blue,
          draw=black] 
          plot coordinates{ 
            (0,0) (1,0.0323834196891192) (2,0.0323834196891192) (3,0.291450777202073) (4,0.420984455958549) (5,0.939119170984456) (6,1.81347150259067) (7,2.16968911917098) (8,1.87823834196891) (9,1.42487046632124) (10,1.58678756476684)
          };
          \addplot[
          ybar,
          bar width=0.04in, 
          bar shift=0.04in,
          fill=red,
          draw=black] 
          plot coordinates{ 
            (0,0) (1,0) (2,0) (3,0) (4,0) (5,0.435540069686411) (6,1.21951219512195) (7,0.958188153310104) (8,0.348432055749129) (9,0.174216027874564) (10,0) (11,0.435540069686411) (12,0.174216027874564) (13,0) (14,0.348432055749129) (15,0.435540069686411)
          };
        
        \end{axis}
    \end{tikzpicture}

    \caption{Normalized histogram of percentage of estimated middle paths being outside of the actual track at a certain distance from the car.}
    \label{fig:hist_oot_dist}
    
    \end{subfigure}
    
     \caption{Histograms to assess the robustness of our system. The yellow and blue bars represent the previous and current implementations on the previous dataset \cite{amz_paper_2019}, respectively. The red bar represents our implementation on \emph{FSG 2019 Autocross} dataset.}
    \label{fig:hist_planning_results}
    
\end{figure}

Fig.~\ref{fig:hist_pred_dist}, verifies the robustness of our system by illustrating the distribution of the output paths according to the path length. It can be observed that when the previous perception data was used, paths were limited to a maximum length of 10~m, and most of the paths generated with this data had a length near the 10~m limit. Using the \emph{FSG 2019 Autocross} data, the maximum path length was increased to  15~m, due to the improvement in our perception range and accuracy. 

Fig.~\ref{fig:hist_oot_dist}, shows that although most predicted paths reach the largest allowable length, some paths went out of the track midway. Generally, the shorter a path is before going out of the track, the lower the buffer for correction as well. Therefore, it is crucial for points on the path that are closer to the car to be estimated correctly. 
Furthermore, the histogram in Fig.~\ref{fig:hist_oot_dist} shows that the current implementation is more robust as the previous boundary estimation approach had a higher percentage of paths going out of the track at distances close to the car. 
Additionally, it can be observed that our improved perception is also a contributing factor, as the value of the red bar is either lower or about the same as compared to the other two bars for distances up to 10~m.

\section{CONCLUSION}
% in the end
\label{sec:conclusion}

This paper presented the perception, mapping and planning components of the autonomous driving system developed by the 2019 AMZ driverless team for the \emph{Autocross} discipline. Our early fusion approach of combining point clouds and images achieved better results than a previous LiDAR-only approach across various distances. In mapping, a local map and global map were generated. Filtering the cone observations by fusing associated past and current cone observations generated the local map. Using only high-certainty observations from the local map, an accurate, globally consistent map was generated via pose-landmark GraphSLAM. Our experiments showed that compared to our previous implementation, a lower RMSE was achieved at the same mapping velocity, and the mapping velocity could be increased greatly without a large sacrifice on accuracy. Lastly, for planning, we presented a Bayesian inference method for estimating the middle path and its corresponding boundaries. We demonstrated higher robustness as a lower fraction of paths that left the track at distances nearer to the car when compared to our previous implementation on the same data set. 
These new methods enable faster driving with similar mapping performance, even at speeds 4 times the previous implementation.

\addtolength{\textheight}{-12cm}   % This command serves to balance the column lengths

%%%%%%%%%%%%%%%%%%%%%%%%%%%%%%%%%%%%%%%%%%%%%%%%%%%%%%%%%%%%%%%%%%%%%%%%%%%%%%%%

%%%%%%%%%%%%%%%%%%%%%%%%%%%%%%%%%%%%%%%%%%%%%%%%%%%%%%%%%%%%%%%%%%%%%%%%%%%%%%%%

%%%%%%%%%%%%%%%%%%%%%%%%%%%%%%%%%%%%%%%%%%%%%%%%%%%%%%%%%%%%%%%%%%%%%%%%%%%%%%%%

\section*{ACKNOWLEDGMENT}

The authors would like to thank the whole AMZ driverless team for their dedication and hard work in developing \emph{pilatus driverless}, and the team's sponsors for their financial and technical support throughout the season. We also wish to express our appreciation for everyone at the Autonomous Systems Lab of ETH Zürich for their supervision and support throughout this project.

%%%%%%%%%%%%%%%%%%%%%%%%%%%%%%%%%%%%%%%%%%%%%%%%%%%%%%%%%%%%%%%%%%%%%%%%%%%%%%%%

%References are important to the reader; therefore, each citation must be complete and correct. If at all possible, references should be commonly available publications.

\begin{acronym}
\acro{FSD}{Formula Student Driverless}
\acro{SLAM}{Simultaneous Localization and Mapping}
\acro{UKF}{Unscented Kalman Filter}
\end{acronym}

\bibliography{Bibliography}{}
\bibliographystyle{IEEEtran}
\end{document}